\documentclass{article}

    \usepackage[final,nonatbib]{neurips_2020}

\usepackage[utf8]{inputenc} 
\usepackage[T1]{fontenc}    
\usepackage{hyperref}       
\usepackage{url}            
\usepackage{booktabs}       
\usepackage{amsfonts}       
\usepackage{nicefrac}       
\usepackage{microtype}      
\usepackage{xcolor}
\usepackage{graphicx}
\usepackage{subfigure}
\usepackage{listings}
\usepackage{multirow}
\usepackage{wrapfig}

\renewcommand\footnotemark{}

\title{COVID-19 India Dataset: Parsing COVID-19 Data in Daily Health Bulletins from States in India}

\author{
    Mayank Agarwal$^{1}$ 
    $\cdot$ \textnormal{\bf Tathagata Chakraborti$^{1}$}
    $\cdot$ \textnormal{\bf Sachin Grover$^{2}$}
    $\cdot$ \textnormal{\bf Arunima Chaudhary$^{1}$}\\[1em]
    $^1$IBM Research\\
    $^2$Arizona State University
}

\begin{document}

\maketitle

\begin{abstract}
While India has been one of the hotspots of COVID-19, 
data about the pandemic from the country has proved to be largely 
inaccessible at scale.
Much of the data exists in unstructured form on the web, and 
limited aspects of such data are available through public APIs 
maintained manually through volunteer effort.
This has proved to be difficult both in terms of ease of access
to detailed data and with regards to the maintenance 
of manual data-keeping over time.
This paper reports on our effort at automating 
the extraction of such data from public health bulletins with the help of
a combination of classical PDF parsers and
state-of-the-art machine learning techniques. 
In this paper, we will describe the automated data-extraction technique,
the nature of the generated data, and exciting avenues of ongoing work.

\vspace{5pt}
{\bf Link} \textcolor{blue}{\url{ibm.biz/covid-data-india}}
\end{abstract}

\section{Introduction}
\label{sec:intro}

Availability of COVID-19 data is crucial for researchers and policymakers 
to understand the pandemic and react to it in real-time. 
However, unlike countries with well-defined data reporting mechanisms, 
pandemic data from India is available either through volunteer-driven initiatives, 
through special access granted by the government, 
or manually collected from daily bulletins 
published by states and cities on their own websites 
or platforms. 

While daily health bulletins from Indian states contain a wealth of data,
they are only available in the unstructured form in PDF documents and images.
On the other hand, volunteer-driven manual data-curation cannot scale
to the volume of data over time. 
For example, one of the most well-known sources of COVID data from India:
\url{covid19india.org}, has manually maintained public APIs for limited data
throughout the pandemic. 
Such approaches, while simultaneously limited in the detail of data made
available, are also unlikely to continue in the long term due to the 
amount of volunteer manual labor required indefinitely.
Although this project originally began anticipating that outcome, 
that eventuality has already come to pass for the aforementioned 
project, for similar reasons outlined in \cite{sunset}.
As such, detailed COVID-19 data from India, in a structured form,
remains inaccessible at scale. 
\cite{plea} notes pleas from researchers in India, earlier this year, for 
the urgent access to detailed COVID data collected by government agencies. 

The aim of this project is to use document and image extraction techniques
to automate the extraction of such data in structured (SQL) form from the 
state-level daily health bulletins; and make this data freely available.
Our target is to automate the data extraction process, so that once the extraction for each state is complete, it requires little to no attention after that (other than responding to changes in the schema).
The role of machine learning here is to make that extraction automated and
robust in coverage and accuracy.
This data goes beyond just daily case and vaccinations numbers to 
comprehensive state-wise metrics such as the hospitalization data, 
age-wise distribution of cases, asymptomatic and symptomatic cases, and even 
case information for individuals in certain states. 

India, one of the most populous countries in the world,
has reported over 33 million confirmed cases of COVID-19 -- 
second only to the United States. 
The massive scale of this data not only provides intriguing research opportunities in data science, document understanding, and NLP for AI researchers but will also help epidemiologists and public policy experts 
to analyze and derive key insights about the pandemic in real-time. 
At the time of this writing, \url{covid19india.org} has also released
possible alternatives going forward once the current APIs are sunset 
next month. These suggestions, detailed here: \cite{operations}, 
also align perfectly
with this current project and give us hope that 
we can continue providing this data, at scale and with much more
detail than ever before. 

\begin{figure}
\begin{center}
\includegraphics[width=\textwidth]{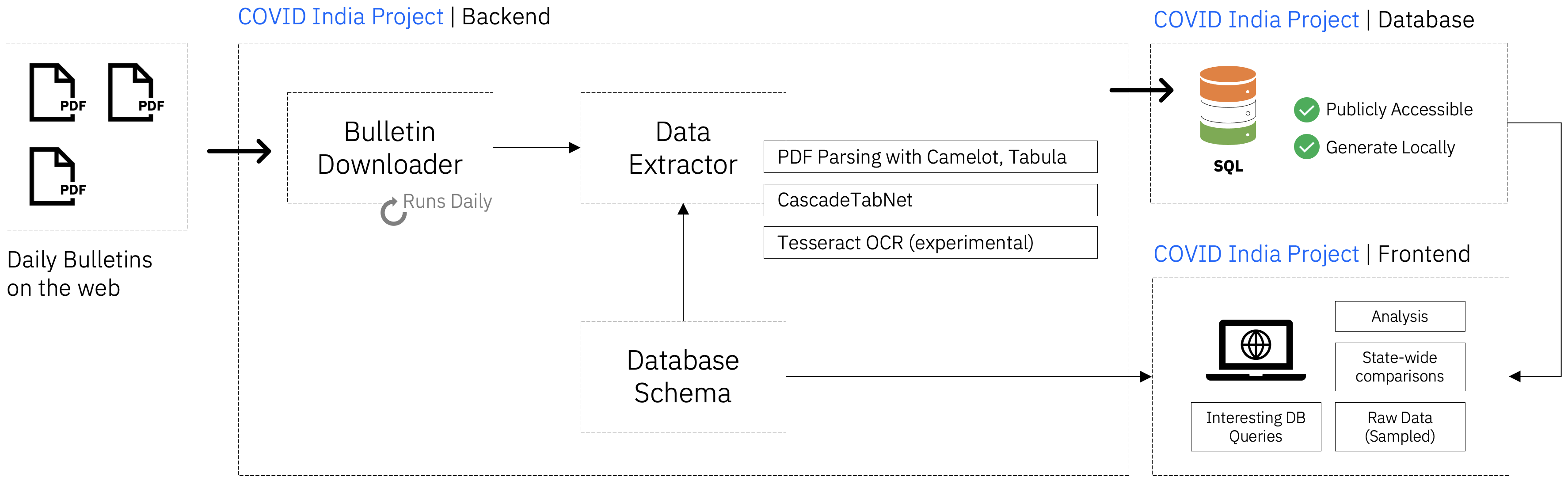}
\end{center}
\caption{Illustration of the data extraction pipeline from daily health bulletins
to an SQL database.}
\label{fig:architecture}
\end{figure}


\section{System Overview}
\label{sec:arch}

We segment the system into 3 major components: (a) the backend which is responsible for extracting data from health bulletins, (b) the database which stores the parsed structured data, and (c) the frontend which displays key analyses extracted from the parsed data. We describe each of these components in greater detail in the following sections. 





\subsection{The Backend}

Since we aim to extract data from health bulletins published by individual states on their respective websites, there is no standard template that is followed across these data sources in terms of where and how the bulletin is published, and what and how information is included in these bulletins. To account for these variations, we modularize the system into the following 3 main components: a) bulletin download, b) datatable definition, and c) data extraction. We provide an overview of the system in Figure \ref{fig:architecture} and look at the three components in greater detail. 
The open-sourced code can be accessed at:
\textcolor{blue}{\url{https://github.com/IBM/covid19-india-data}}.

\subsubsection{Bulletin download}

The bulletin download procedure downloads the bulletins from the respective state websites to the local storage while maintaining the dates already processed. We use the BeautifulSoup \footnote{\url{https://www.crummy.com/software/BeautifulSoup/}} library to parse the state websites and identify bulletin links and dates for download.

\subsubsection{Datatable definitions}

Since each state provides different information, we define table 
schemas for each state by manually investigating the bulletin (done once per state). We then use the free open-source SQLite \footnote{\url{https://www.sqlite.org/index.html}} database to interface with the data extractor and store the data.

\subsubsection{Data extractor}

States typically provide the bulletins in the form of
PDF
documents. To extract information from them, we use a combination of classical PDF parsers and state of the art Machine Learning based extraction techniques: 

\vspace{-5pt}
\paragraph{Classical PDF parsing:} Since a substantial amount of information in the bulletins are in the form of data tables, we use the Tabula\footnote{\url{https://tabula.technology/}} and the 
Camelot\footnote{\url{https://camelot-py.readthedocs.io/en/master/}} Python libraries to extract these tables in the form of python data structures. While these libraries cover a lot of use cases, they do fail in certain edge case scenarios.

\vspace{-5pt}
\paragraph{Deep-learning augmented PDF parsing:} Libraries extracting data tables from PDF
typically use either the Lattice
or the Stream \cite{Elomaa2013ANSSINA} based method of detecting table boundaries and
inferring table structure. While these heuristics works great for most cases, 
for cases where tables are either not well separated or are spread wide, 
they fail to correctly separate tables with each other, and group all the tables together.
To correct for such errors, we utilize CascadeTabNet \cite{cascadetabnet2020}, a state-of-the-art convolutional neural network that identifies table regions and structure. We use the detected table boundaries to parse for tables in areas of the PDF, thereby increasing the parsing accuracy. We show an example of performance gain we get from this approach in Appendix \ref{sec:cascadetabnet-example}.

\vspace{-5pt}
\paragraph{Data extraction from images:} While a majority of information provided in health bulletins is in the form of textual tables, some information is provided as images of tabular data. This information cannot be processed through the aforementioned techniques, and requires Optical Character Recognition (OCR) to extract data from. We employ the Tesseract OCR engine \cite{smith2007overview} to read and extract tabular data provided as images. In Appendix \ref{sec:ocr-example}, we provide an example of a bulletin parsed through Tesseract OCR. The detected text is overlayed in the green boxes. 
Note that this is an experimental feature and we are actively working on assessing and improving its efficacy.


To process information for a state, a separate data extractor routine is used, which has access to all the three aforementioned APIs. Depending on the format of the particular bulletin, we utilize a combination of the three techniques to extract information.

\subsection{The Frontend}
\label{sec:frontend}

The frontend or landing page for the project is generated automatically 
from the database schema and provides easy access to 
1) the raw data (sampled at an appropriate rate to be loaded on the browser); and
2) pages for highlights and analysis based on SQL queries (such as those 
described in Section \ref{sec:prelim-analysis}).

\subsection{The Database}
\label{sec:data-description}

The system described above runs daily and produces a SQL database that 
is publicly available for download.
However, one can also use the source code to generate data customized 
with their own parameters, and deploy into their local systems.


\vspace{-5pt}
\paragraph{Current Status:}
At the time of writing, 
we have completely indexed information from seven 
major Indian states, covering a population of over
382 million people or roughly 28.67\% of India's 
population. Additionally, we're in the final stages of
integrating 5 new states, covering an additional 271.5 million
people in the database, for a total coverage of 653.5 million people.
In Appendix \ref{sec:data-comparison}, we provide an overview of the categories
of information available in our database, and contrast it with
the information in the covid19india.org database.

\section{Preliminary Analysis}
\label{sec:prelim-analysis}

In this section, we perform some preliminary analysis on the data collected from the health bulletins of Delhi and West Bengal. We would like to emphasize that some of these analyses (to the best of our knowledge) are the first such analyses available for the two states. However, these are still preliminary but provide an insight into the power of such data available to researchers interested in the subject.

\begin{figure*}[!t]
\centering

\subfigure[Weekly CFR]{\includegraphics[width=0.24\textwidth]{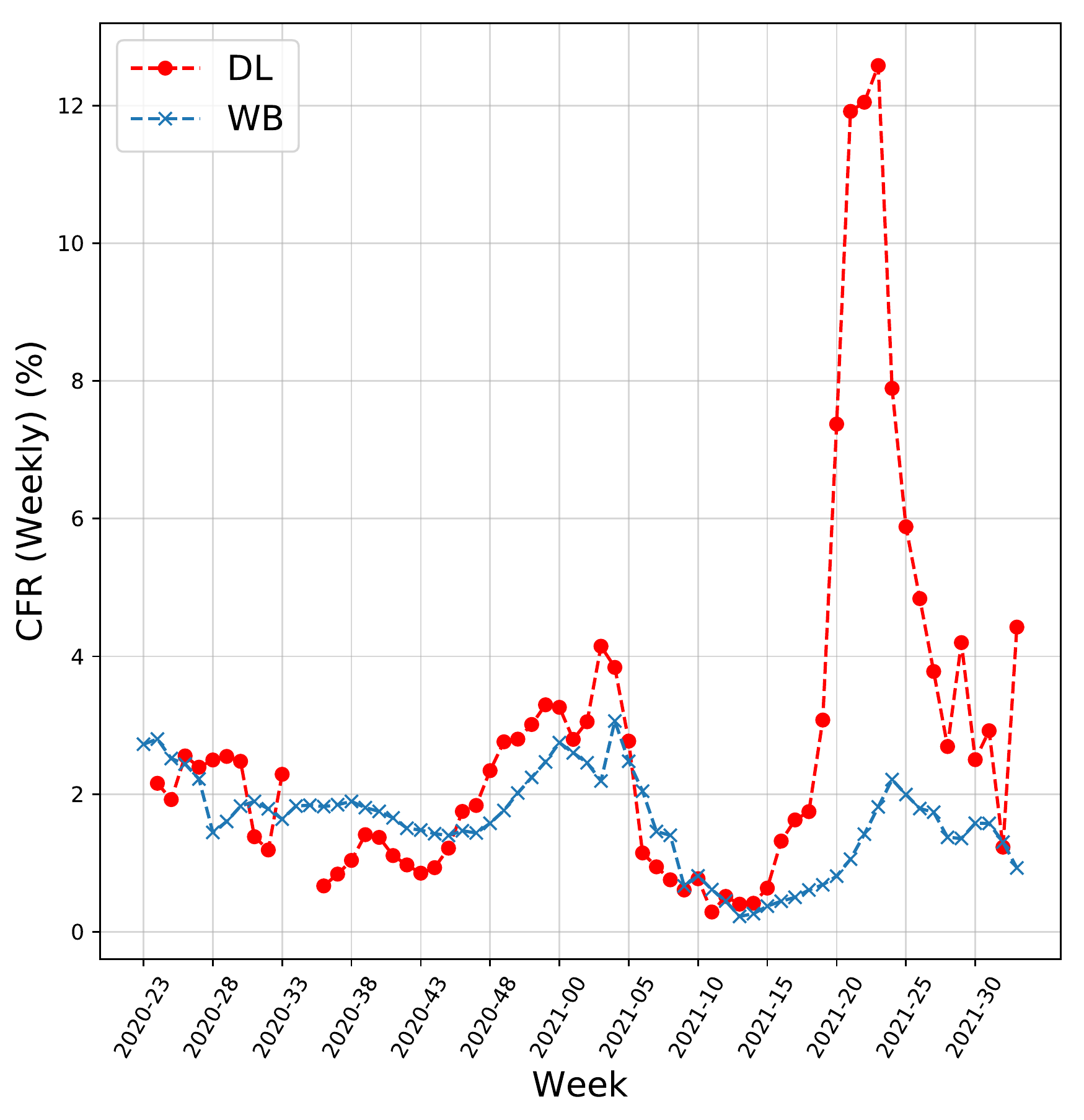}} \hfill
\subfigure[RTPCR tests (DL)]{\includegraphics[width=0.24\textwidth]{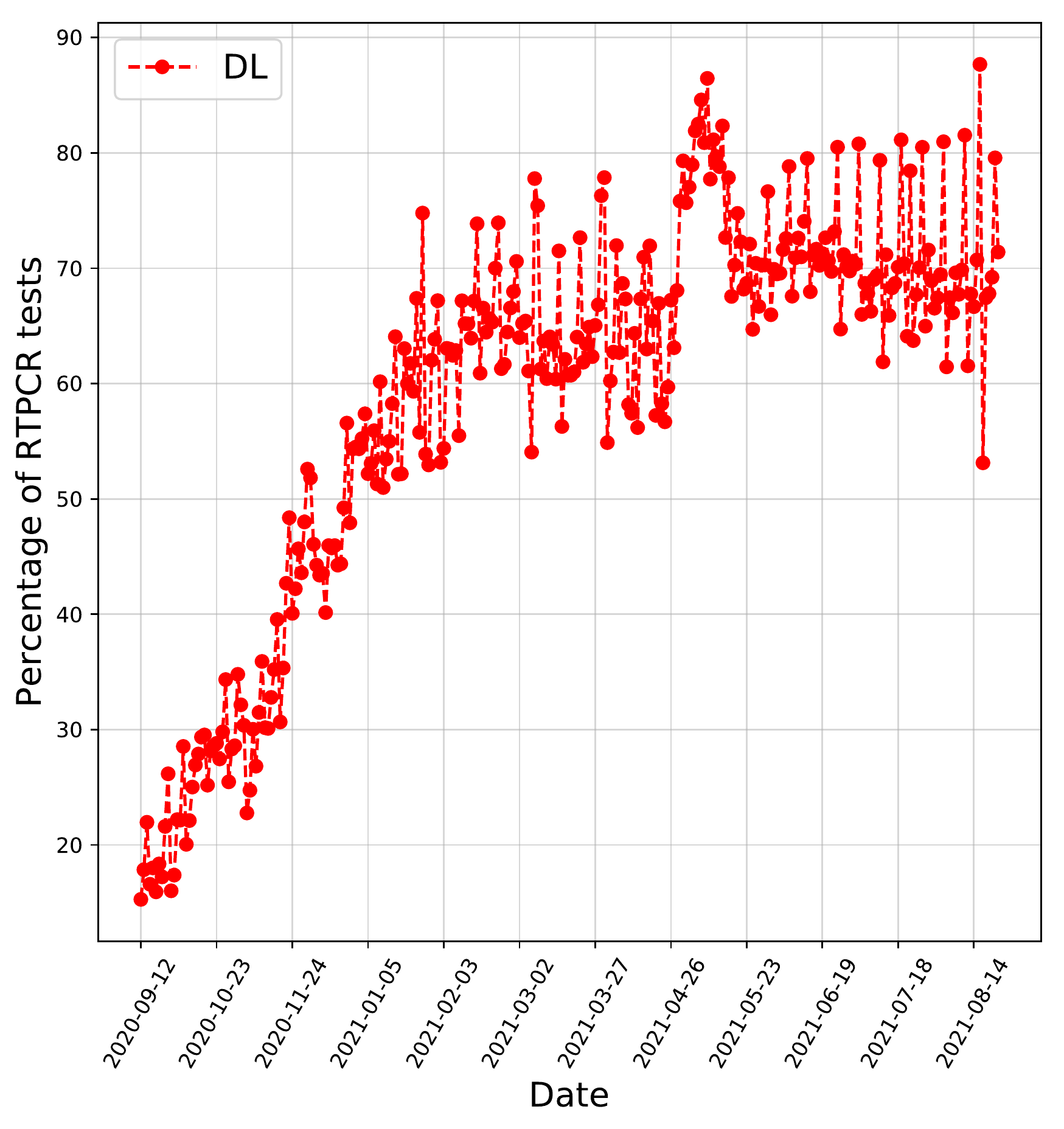}} \hfill
\subfigure[Bed occupancy]{\includegraphics[width=0.24\textwidth]{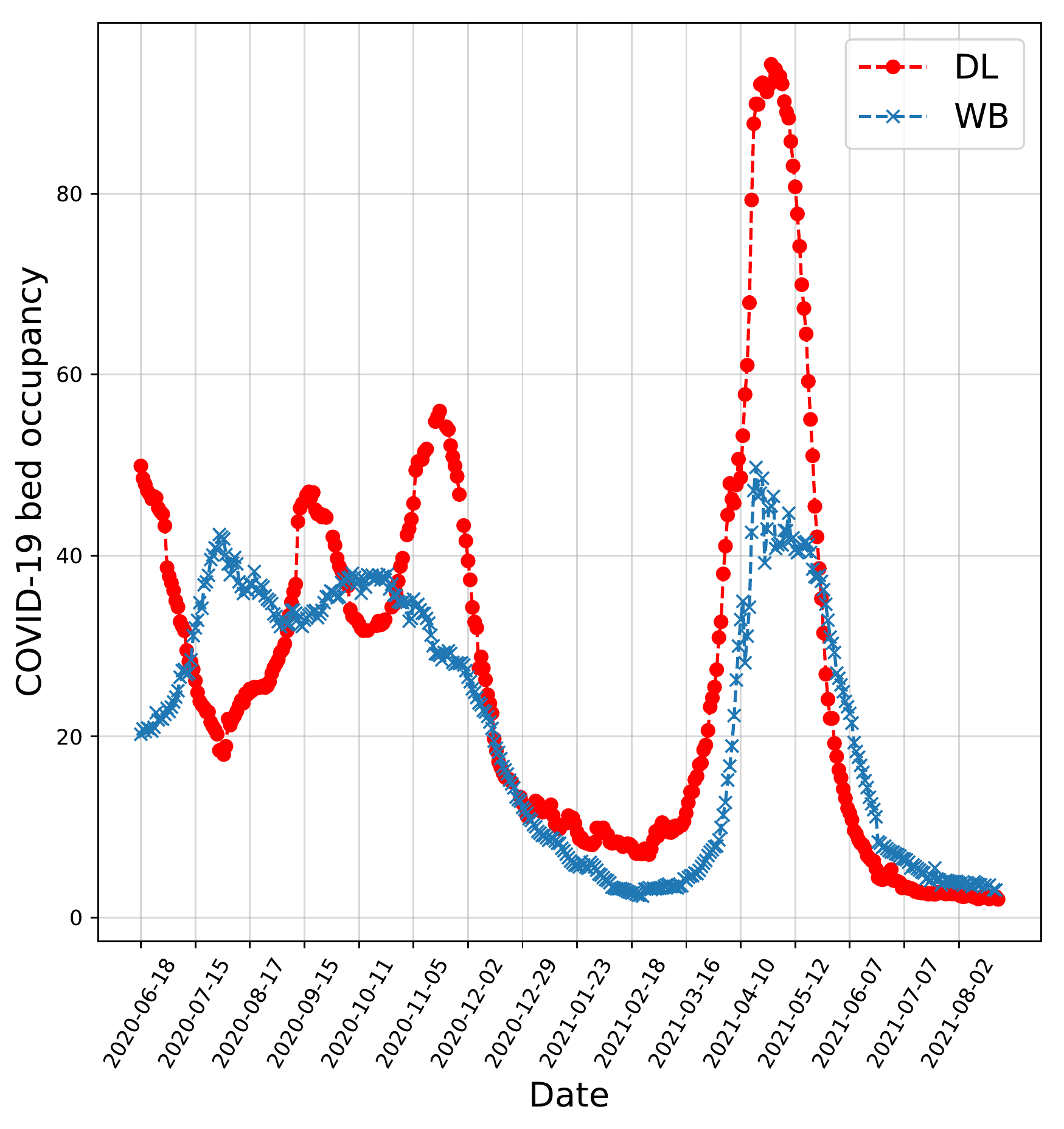}} \hfill
\subfigure[Hospitalization \%-age]{\includegraphics[width=0.24\textwidth]{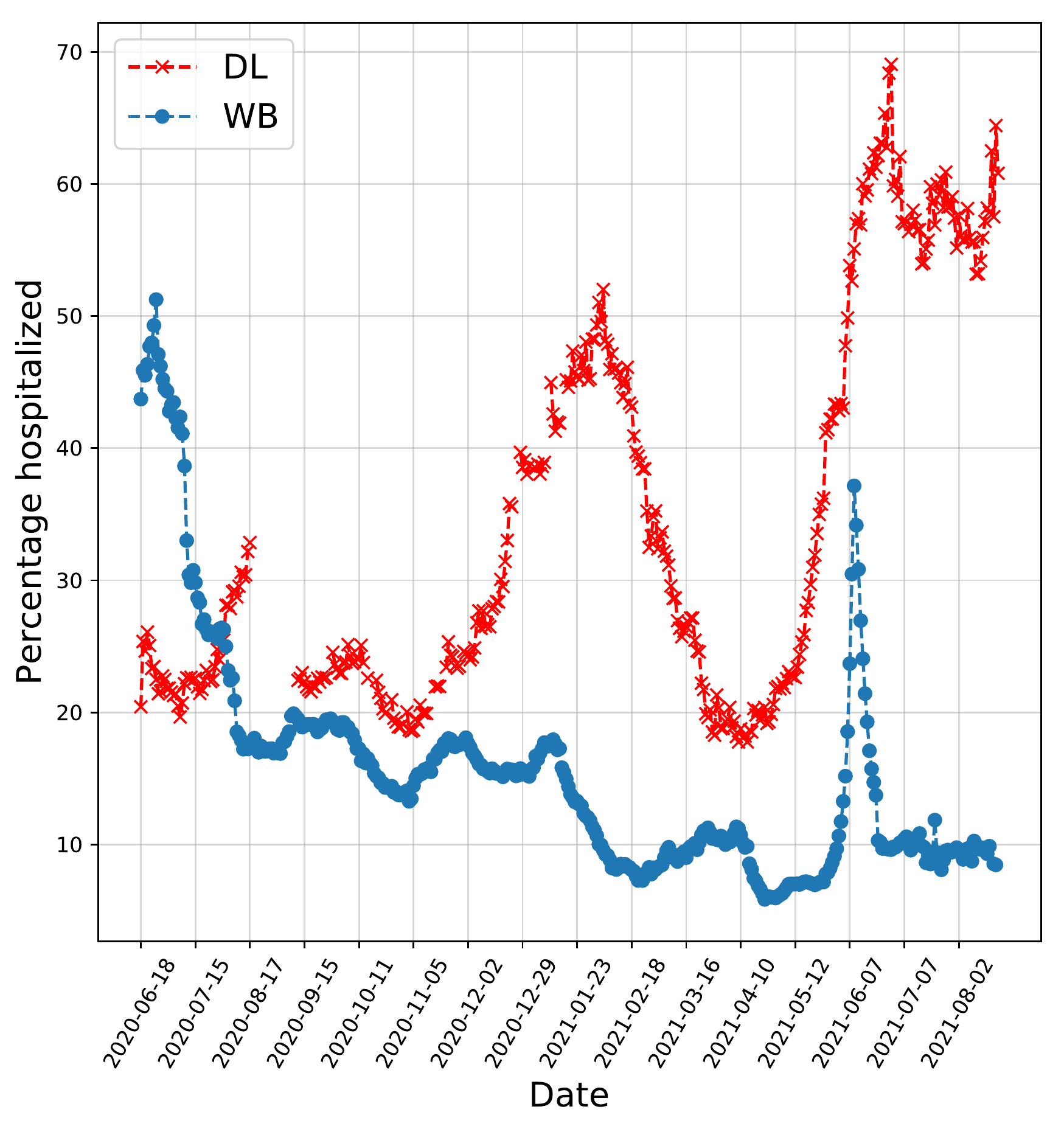}}

\caption{Preliminary analysis illustrating the depth of data available from the daily health bulletins.}
\label{fig:DL_analysis}
\end{figure*}

\subsection{Weekly Case Fatality Rate (CFR)}
\label{sec:weekly-cfr}

India has seen two major waves of COVID-19, with the second wave fuelled primarily by the Delta variant \cite{yang2021covid} being more deadly than the first \cite{budhiraja2021differentials, gupta2021clinical}. We aim to understand the difference between the two waves by computing the Weekly Case Fatality Rate as the ratio of total fatalities to total newly confirmed cases in a particular week. The charts for Delhi and West Bengal are presented in Figure \ref{fig:DL_analysis}. While the weekly CFR for the first wave seems to be comparable for the two states, there appears to be a stark difference in the numbers for the second wave.

\subsection{Percentage of RT-PCR tests}
\label{sec:rtpcr}

Currently, India uses the reverse-transcriptase polymerase-chain-reaction (RT-PCR) tests and the Rapid Antigen Tests (RATs) to detect COVID-19 cases. While RT-PCR tests are highly accurate and are considered gold-standard tests for detecting COVID-19 \cite{brihn2021diagnostic}, they are more expensive and time-consuming than the less accurate RATs. 
While the official advisory is to prefer RT-PCRs over RATs \cite{icmr_testing_adv}, there exists a discrepancy in how the two testing methods are used \cite{cherian2021optimizing} and how this ratio affects the reported case results \cite{chatterjee2020india}.
The state government of Delhi has in the past been called out for over-reliance on RATs as opposed to the preferred RT-PCR tests \cite{sirur_2020}. Following this criticism, the government increased the share of RT-PCR tests. We compute this ratio of RT-PCR tests to total tests conducted in the state (Figure \ref{fig:DL_analysis}). 
As is evident, in 2020, less than 50\% of the total tests conducted in the state were RT-PCR tests. However, starting 2021, and especially during the second wave of COVID-19 in India, this ratio increased to over 70\%.

\subsection{COVID-19 bed occupancy}
\label{sec:bed-occupancy}

Both DL and WB report the dedicated COVID-19 hospital infrastructure and occupancy information in their bulletins. Using these numbers, we compute the COVID-19 bed occupancy as the ratio of occupied beds to total (Figure \ref{fig:DL_analysis}). Similar to the results in Section \ref{sec:weekly-cfr}, bed occupancy for Delhi shows a steep increase -- reaching about 90\% occupancy -- during the second wave, while the occupancy for West Bengal does not show any significant difference during the two waves.

\subsection{Hospitalization percentage}
\label{sec:hospitalization}

To treat COVID-19 patients, India adopted a two-pronged strategy of hospitalization along with home isolation, where patients with a mild case of COVID-19 were advised home isolation whereas hospitals were reserved for patients with more severe cases of COVID-19 \cite{varghese2020covid, bhardwaj2021analysis}. 
We compute the hospitalization percentage as the ratio of the number of occupied hospital beds to the number of active cases.
This is an estimate of how many of the currently active COVID-19 patients are in hospitals versus home isolation (Figure \ref{fig:DL_analysis}). The peaks we see for the two states relate to time periods after the respective wave has subsided
, the minima and the subsequent rise in hospitalization relate to the onset of the particular wave. 

\section{Future work}
\label{sec:future-work}

The primary aim of this project is to extract as much information about the pandemic as possible from public sources so that this data can be made accessible in an easy and structured form to researchers who can utilize such data (from one of the most populous and heavily COVID-affected countries in the world) in their research. We foresee two main areas of future work for this project:

\begin{enumerate}
\item In the immediate future, we aim to integrate information for all Indian states into the dataset. Additionally, the project currently relies on health bulletins alone to extract the data. There are other platforms where the authorities release data, such as Twitter and Government APIs \cite{covid19org_sources}. We hope to integrate these additional sources of information into the dataset.


\item We anticipate this data to be helpful in validating or extending models developed for other countries \cite{friedman2021predictive, borchering2021modeling},
developing pandemic models which integrate additional variables available in our dataset \cite{hethcote2000mathematics, agrawal2021sutra, adiga2020mathematical, Bhaduri2020ExtendingTS}, and understanding other aspects of the pandemic \cite{Ray2020PredictionsRO, Ghosh2020InterstateTP}.
\end{enumerate}

\begin{ack}
We would like to thank all our open source contributors, in addition to those who
have joined as as co-authors of this paper, for their amazing contributions to 
this project and this dataset. In particular, we thank Sushovan De (Google) for 
helping us extending the dataset to the Indian state of Karnataka.
\end{ack}

\bibliographystyle{plain}
\bibliography{covid19indiadata}

\begin{thebibliography}{10}

\bibitem{adiga2020mathematical}
Aniruddha Adiga, Devdatt Dubhashi, Bryan Lewis, Madhav Marathe, Srinivasan
  Venkatramanan, and Anil Vullikanti.
\newblock {Mathematical models for COVID-19 pandemic: a comparative analysis}.
\newblock {\em Journal of the Indian Institute of Science}, pages 1--15, 2020.

\bibitem{agrawal2021sutra}
Manindra Agrawal, Madhuri Kanitkar, and Mathukumalli Vidyasagar.
\newblock {SUTRA: An approach to modelling pandemics with asymptomatic
  patients, and applications to COVID-19}.
\newblock {\em arXiv preprint arXiv:2101.09158}, 2021.

\bibitem{Bhaduri2020ExtendingTS}
R.~Bhaduri, R.~Kundu, S.~Purkayastha, M.~Kleinsasser, L.~Beesley, and
  B.~Mukherjee.
\newblock {Extending the Susceptible-Exposed-Infected-Removed (SEIR) model to
  handle the high false negative rate and symptom-based administration of
  COVID-19 diagnostic tests: SEIR-fansy}.
\newblock {\em medRxiv}, 2020.

\bibitem{bhardwaj2021analysis}
Pankaj Bhardwaj, Nitin~Kumar Joshi, Manoj~Kumar Gupta, Akhil~Dhanesh Goel,
  Suman Saurabh, Jaykaran Charan, Prakash Rajpurohit, Suresh Ola, Pritam Singh,
  Sunil Bisht, et~al.
\newblock {Analysis of Facility and Home Isolation Strategies in COVID 19
  Pandemic: Evidences from Jodhpur, India}.
\newblock {\em Infection and Drug Resistance}, 14:2233, 2021.

\bibitem{borchering2021modeling}
Rebecca~K Borchering, C{\'e}cile Viboud, Emily Howerton, Claire~P Smith, Shaun
  Truelove, Michael~C Runge, Nicholas~G Reich, Lucie Contamin, John Levander,
  Jessica Salerno, et~al.
\newblock {Modeling of future COVID-19 cases, hospitalizations, and deaths, by
  vaccination rates and nonpharmaceutical intervention scenarios -- United
  States, April--September 2021}.
\newblock {\em Morbidity and Mortality Weekly Report}, 70(19):719, 2021.

\bibitem{brihn2021diagnostic}
Auguste Brihn, Jamie Chang, Kelsey OYong, Sharon Balter, Dawn Terashita, Zach
  Rubin, and Nava Yeganeh.
\newblock {Diagnostic Performance of an Antigen Test with RT-PCR for the
  Detection of SARS-CoV-2 in a Hospital Setting—Los Angeles County,
  California, June--August 2020}.
\newblock {\em Morbidity and Mortality Weekly Report}, 70(19):702, 2021.

\bibitem{budhiraja2021differentials}
Sandeep Budhiraja, Abhaya Indrayan, Mona Aggarwal, Vinita Jha, Dinesh Jain,
  Bansidhar Tarai, Poonam Das, Bharat Aggarwal, RS~Mishra, Supriya Bali, et~al.
\newblock {Differentials in the characteristics of COVID-19 cases in Wave-1 and
  Wave-2 admitted to a network of hospitals in North India}.
\newblock {\em medRxiv}, 2021.

\bibitem{chatterjee2020india}
Patralekha Chatterjee.
\newblock {Is India missing COVID-19 deaths?}
\newblock {\em The Lancet}, 396(10252):657, 2020.

\bibitem{cherian2021optimizing}
Philip Cherian, Sandeep Krishna, and Gautam~I Menon.
\newblock {Optimizing Testing for COVID-19 in India}.
\newblock {\em medRxiv}, pages 2020--12, 2021.

\bibitem{covid19org_sources}
covid19india.
\newblock {Hornbill}.
\newblock \url{https://blog.covid19india.org/2020/06/15/hornbill/}, 2020.

\bibitem{operations}
{covid19india}.
\newblock {Operations}.
\newblock \url{https://blog.covid19india.org/2021/08/24/operations/}, August
  2021.

\bibitem{sunset}
{covid19india}.
\newblock {When the curtains come down}.
\newblock \url{https://blog.covid19india.org/2021/08/07/end/}, August 2021.

\bibitem{Elomaa2013ANSSINA}
Tapio Elomaa.
\newblock Anssi nurminen algorithmic extraction of data in tables in pdf
  documents.
\newblock 2013.

\bibitem{friedman2021predictive}
Joseph Friedman, Patrick Liu, Christopher~E Troeger, Austin Carter, Robert~C
  Reiner, Ryan~M Barber, James Collins, Stephen~S Lim, David~M Pigott, Theo
  Vos, et~al.
\newblock {Predictive performance of international COVID-19 mortality
  forecasting models}.
\newblock {\em Nature Communications}, 12(1):1--13, 2021.

\bibitem{Ghosh2020InterstateTP}
K.~Ghosh, Nairita Sengupta, Dipanwita Manna, and S.~De.
\newblock {Inter-state transmission potential and vulnerability of COVID-19 in
  India}.
\newblock {\em Progress in Disaster Science}, 7:100114 -- 100114, 2020.

\bibitem{gupta2021clinical}
Nivedita Gupta, Harmanmeet Kaur, Pragya Yadav, Labanya Mukhopadhyay, Rima~R
  Sahay, Abhinendra Kumar, Dimpal~A Nyayanit, Anita~M Shete, Savita Patil,
  Triparna~Dutta Majumdar, et~al.
\newblock {Clinical characterization and Genomic analysis of COVID-19
  breakthrough infections during second wave in different states of India}.
\newblock {\em medRxiv}, 2021.

\bibitem{hethcote2000mathematics}
Herbert~W Hethcote.
\newblock The mathematics of infectious diseases.
\newblock {\em SIAM review}, 42(4):599--653, 2000.

\bibitem{icmr_testing_adv}
{Indian Council of Medical Research}.
\newblock {Advisory on Strategy for COVID-19 Testing in India}.
\newblock
  \url{https://www.icmr.gov.in/pdf/covid/strategy/Testing_Strategy_v6_04092020.pdf},
  2020.

\bibitem{cascadetabnet2020}
Devashish Prasad, Ayan Gadpal, Kshitij Kapadni, Manish Visave, and Kavita
  Sultanpure.
\newblock Cascadetabnet: An approach for end to end table detection and
  structure recognition from image-based documents, 2020.

\bibitem{plea}
{Priyanka Pulla}.
\newblock {``There are so many hurdles.'' Indian scientists plead with
  government to unlock COVID-19 data}.
\newblock
  \url{https://www.science.org/news/2021/05/there-are-so-many-hurdles-indian-scientists-plead-government-unlock-covid-19-data},
  May 2021.
\newblock Science.

\bibitem{Ray2020PredictionsRO}
D.~Ray, M.~Salvatore, R.~Bhattacharyya, Lili Wang, Jiacong Du, Shariq Mohammed,
  S.~Purkayastha, Aritra Halder, Alexander Rix, D.~Barker, M.~Kleinsasser,
  Yiwang Zhou, Debraj Bose, Peter X.~K. Song, Mousumi Banerjee,
  V.~Baladandayuthapani, P.~Ghosh, and B.~Mukherjee.
\newblock {Predictions, role of interventions and effects of a historic
  national lockdown in India's response to the COVID-19 pandemic: data science
  call to arms}.
\newblock {\em Harvard Data Science Review}, 2020 Suppl 1, 2020.

\bibitem{sirur_2020}
Simrin Sirur.
\newblock {It isn't just Delhi. Kerala, Bihar \& UP also conduct more than 50\%
  rapid antigen tests}.
\newblock
  \url{https://theprint.in/health/it-isnt-just-delhi-kerala-bihar-up-also-conduct-more-than-50-rapid-antigen-tests/550255/},
  Nov 2020.
\newblock ThePrint.

\bibitem{smith2007overview}
Ray Smith.
\newblock An overview of the tesseract ocr engine.
\newblock In {\em Ninth international conference on document analysis and
  recognition (ICDAR 2007)}, volume~2, pages 629--633. IEEE, 2007.

\bibitem{varghese2020covid}
George~M Varghese and Rebecca John.
\newblock {COVID-19 in India: Moving from containment to mitigation}.
\newblock {\em The Indian journal of medical research}, 151(2-3):136, 2020.

\bibitem{yang2021covid}
Wan Yang and Jeffrey Shaman.
\newblock {COVID-19 pandemic dynamics in India and impact of the SARS-CoV-2
  Delta (B. 1.617. 2) variant}.
\newblock {\em medRxiv}, 2021.

\end{thebibliography}

\appendix

\section{Appendix}

\subsection{Dataset characteristics}
\label{sec:data-comparison}

In Table \ref{tab:db-info}, we present the different attributes that are available in our dataset, and contrast it with the popular \texttt{covid19india.org} dataset. While \texttt{covid19india.org} contains the Case, Testing, and Vaccination information for all states, we include additional features, such as, Hospital infrastructure and hospitalization statistics, Individual fatality data, Age and gender distribution of cases, and Mental Health counselling among others.

\begin{table}[h!]
\centering
\resizebox{\linewidth}{!}{%
\begin{tabular}{|c|c|ccccccc|}
\hline
\textbf{Dataset ($\rightarrow$)} & \multirow{2}{*}{covid19india.org} & \multicolumn{7}{c|}{Ours} \\ \cline{1-1} \cline{3-9} 
\textbf{Category ($\downarrow$)} &  & \multicolumn{1}{c|}{DL} & \multicolumn{1}{c|}{HR} & \multicolumn{1}{c|}{KA} & \multicolumn{1}{c|}{MH} & \multicolumn{1}{c|}{TG} & \multicolumn{1}{c|}{UK} & WB \\ \hline
Case information & Y & \multicolumn{1}{c|}{Y} & \multicolumn{1}{c|}{Y} & \multicolumn{1}{c|}{Y} & \multicolumn{1}{c|}{Y} & \multicolumn{1}{c|}{Y} & \multicolumn{1}{c|}{Y} & Y \\ \hline
Testing & Y & \multicolumn{1}{c|}{Y} & \multicolumn{1}{c|}{Y} & \multicolumn{1}{c|}{Y} & \multicolumn{1}{c|}{Y} & \multicolumn{1}{c|}{Y} & \multicolumn{1}{c|}{Y} & Y \\ \hline
Vaccination & Y & \multicolumn{1}{c|}{Y} & \multicolumn{1}{c|}{Y} & \multicolumn{1}{c|}{Y} & \multicolumn{1}{c|}{-} & \multicolumn{1}{c|}{-} & \multicolumn{1}{c|}{Y} & Y \\ \hline
Hospitalization & - & \multicolumn{1}{c|}{Y} & \multicolumn{1}{c|}{Y} & \multicolumn{1}{c|}{-} & \multicolumn{1}{c|}{-} & \multicolumn{1}{c|}{-} & \multicolumn{1}{c|}{-} & Y \\ \hline
Individual fatalities & - & \multicolumn{1}{c|}{-} & \multicolumn{1}{c|}{-} & \multicolumn{1}{c|}{Y} & \multicolumn{1}{c|}{-} & \multicolumn{1}{c|}{-} & \multicolumn{1}{c|}{-} & - \\ \hline
Age/Gender distribution & - & \multicolumn{1}{c|}{-} & \multicolumn{1}{c|}{-} & \multicolumn{1}{c|}{-} & \multicolumn{1}{c|}{-} & \multicolumn{1}{c|}{Y} & \multicolumn{1}{c|}{-} & - \\ \hline
Mental health counselling & - & \multicolumn{1}{c|}{-} & \multicolumn{1}{c|}{-} & \multicolumn{1}{c|}{-} & \multicolumn{1}{c|}{-} & \multicolumn{1}{c|}{-} & \multicolumn{1}{c|}{-} & Y \\ \hline
\end{tabular}
}
\vspace{5pt}
\caption{Summary of information available for individual states in our database. For reference, we list the information available in the covid19india.org database.}
\label{tab:db-info}
\end{table}

\subsection{Deep-learning augmented PDF parsing}
\label{sec:cascadetabnet-example}

\begin{figure*}[!th]

    \centering
    
        \subfigure[Bulletin page]{\includegraphics[width=0.31\textwidth]{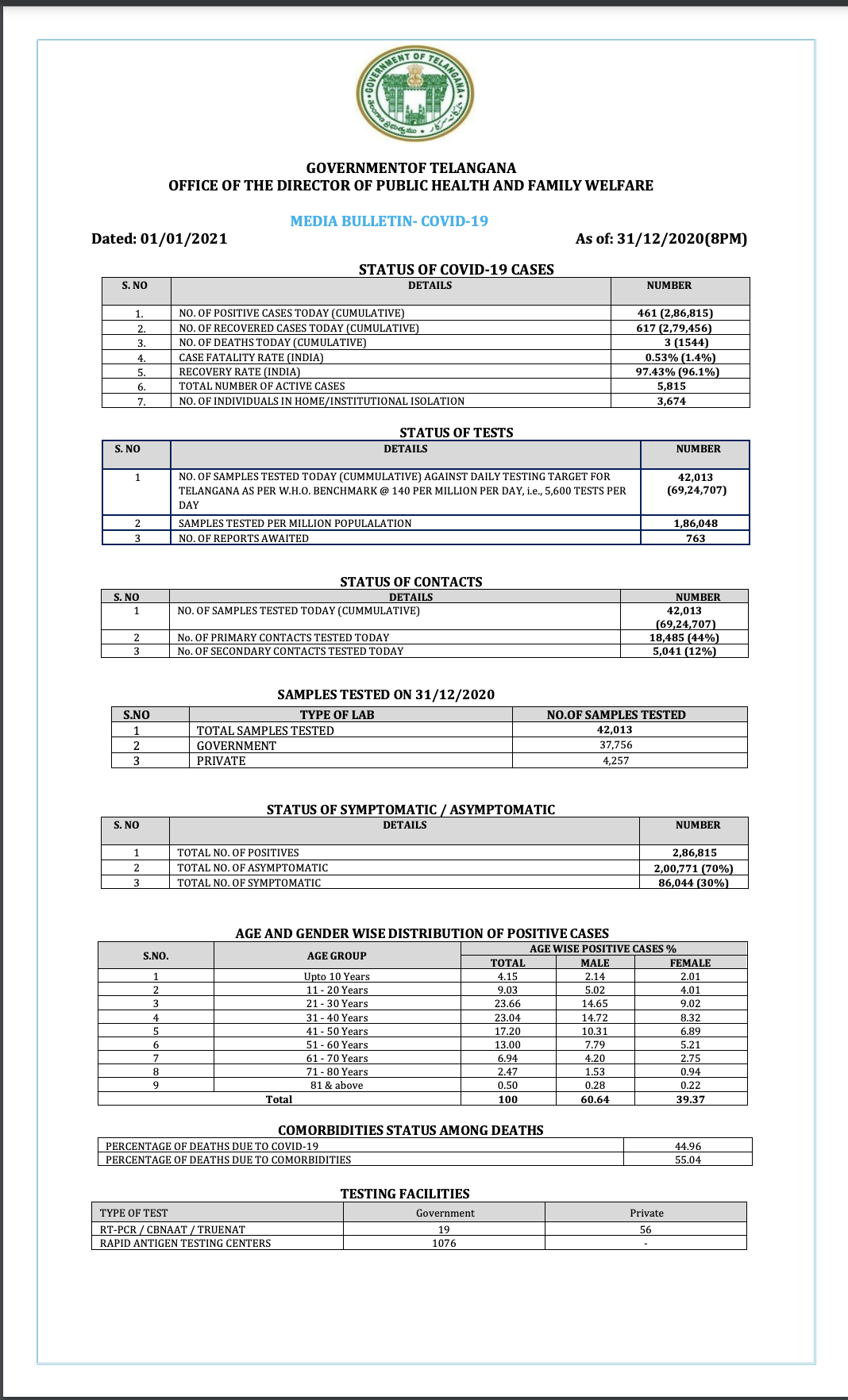}} 
        \subfigure[Native table detection]{\includegraphics[width=0.29\textwidth]{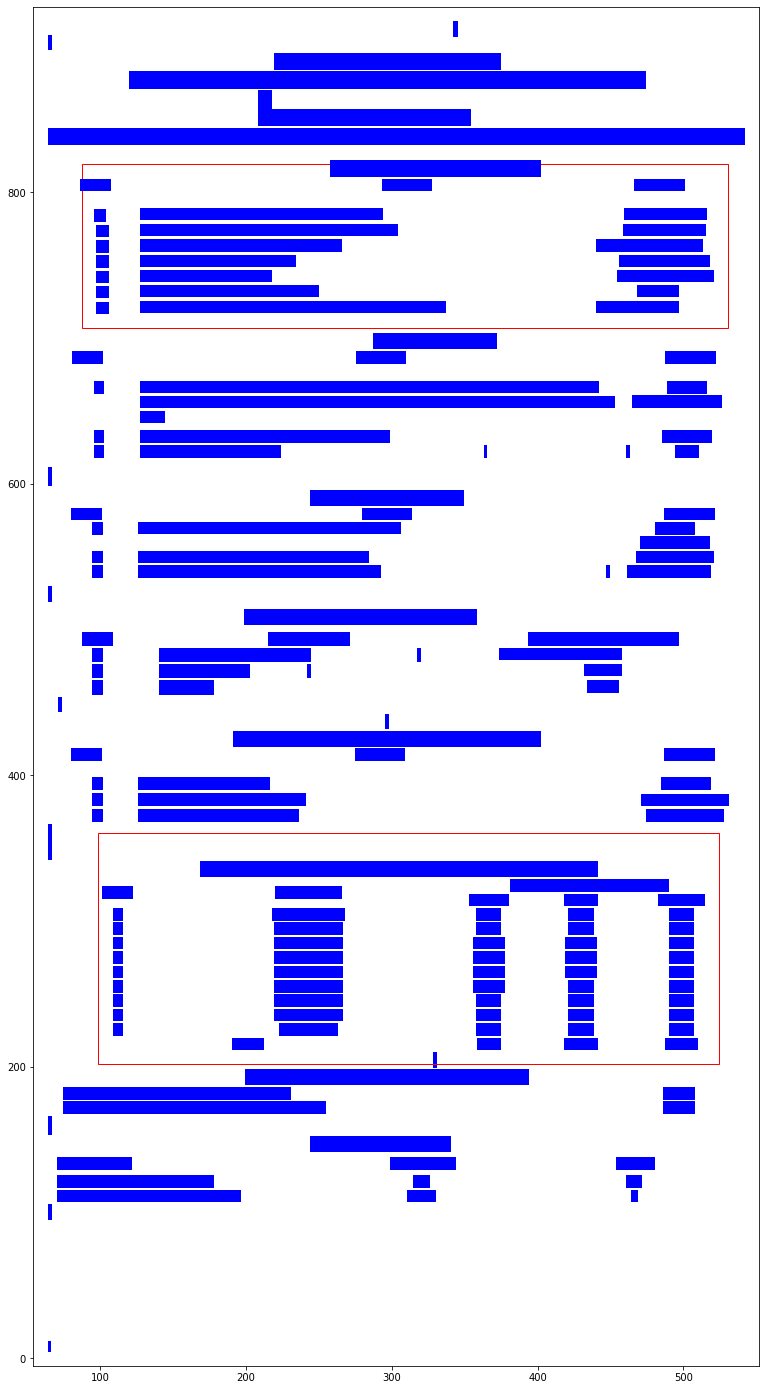}} 
        \subfigure[CascadeTabNet enhanced detection]{\includegraphics[width=0.325\textwidth]{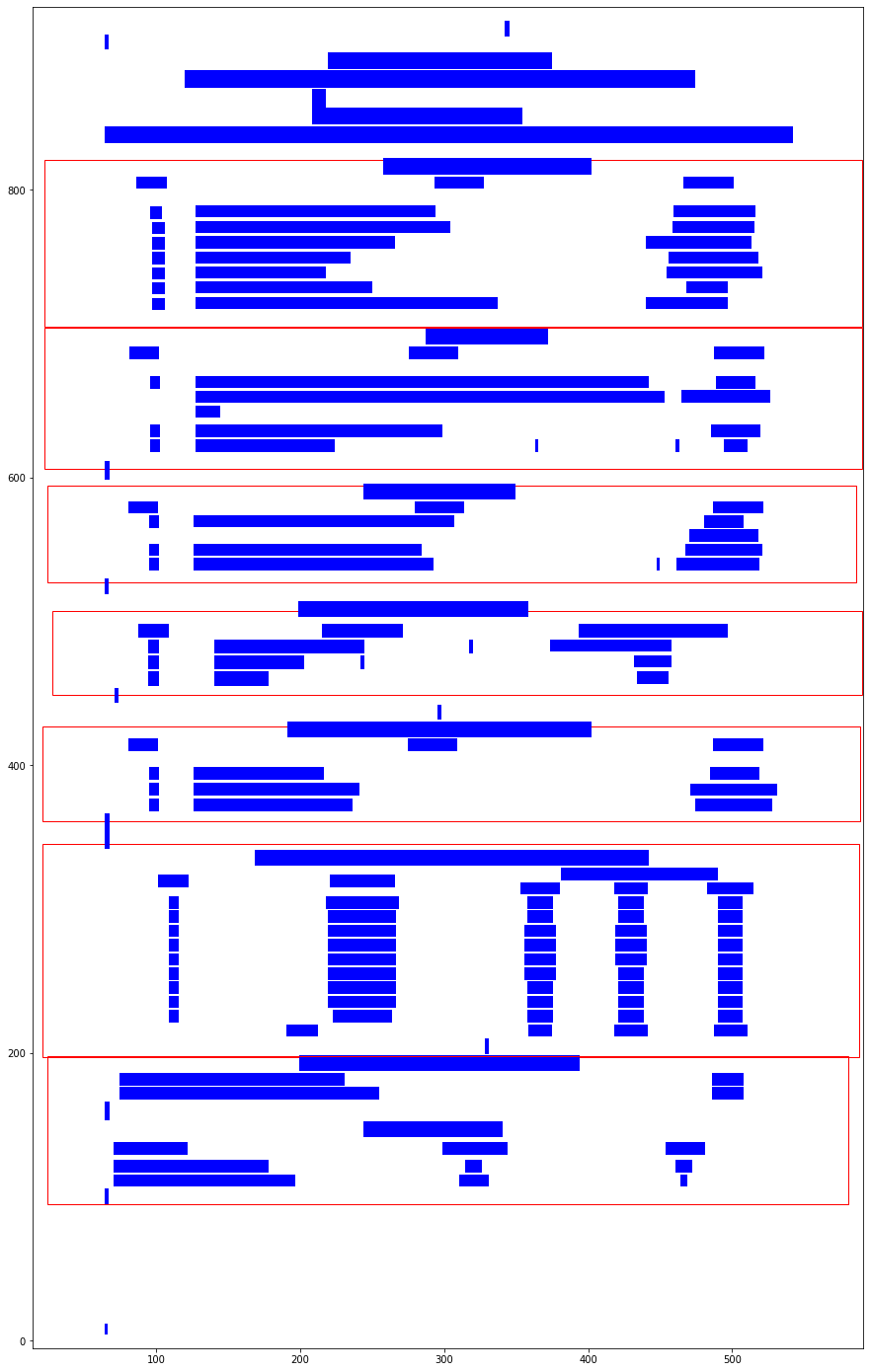}}
    \caption{Table extraction from a state health bulletin using classical PDF parsing and CascadeTabNet enhanced parsing. There are eight tables in the bulletin page (see (a)), and while classical parsing can only detect two tables due to insufficient separation between the tables, CascadeTabNet improves this detection significantly by extracting seven tables but missing one.}
    \label{fig:table-detection}
\end{figure*}

In Figure \ref{fig:table-detection}, we show an example of CascadeTabNet enhanced table extraction. While native table extraction fails to correctly identify the tables due to insufficient separation between the tables, CascadeTabNet correctly detects seven out of the eight tables on the page. Using the table regions identified by CascadeTabNet, we are able to improve the table extraction accuracy for bulletins.

\subsection{OCR based data extraction from images}
\label{sec:ocr-example}

\begin{figure}[!ht]
\centering
\includegraphics[width=0.8\linewidth]{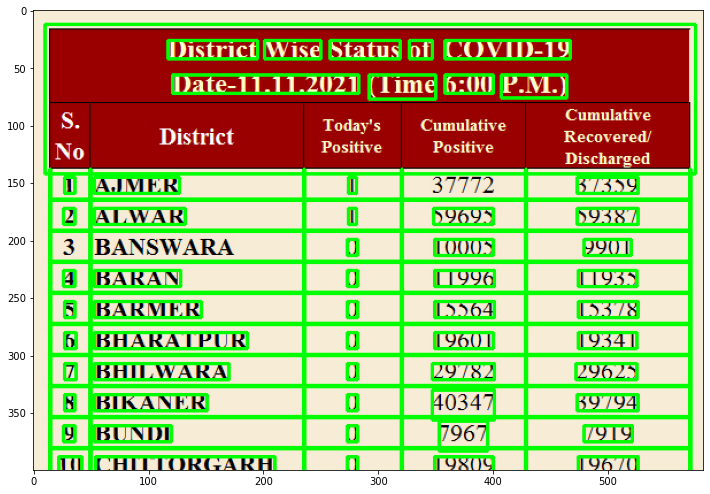}
\caption{State bulletin sample providing tabular data in the form of an image. We use Tesseract OCR to extract data from the image (green bounding boxes). However, the OCR engine fails to extract all the information correctly, for instance, it fails to identify the table header.}
\label{fig:ocr}
\end{figure}

In Figure \ref{fig:ocr}, we show an example of data table provided in the form of an image. Standard table extraction tools do not support extracting data from such format, and therefore we utilize Optical Character Recognition (OCR) for data extraction from such formats. In this figure, we show the detected text and bounding boxes around them. As is evident, this technique fails to identify certain text, such as the header of the table, and certain numbers from the table itself. This is currently an experimental feature, and we're actively working on assessing and improving its efficacy.

\end{document}